\documentclass[lettersize,journal]{IEEEtran}
\usepackage{amsmath,amsfonts}
\usepackage{algorithmic}
\usepackage{algorithm}
\usepackage{array}
\usepackage[caption=false,font=normalsize,labelfont=sf,textfont=sf]{subfig}
\usepackage{textcomp}
\usepackage{stfloats}
\usepackage{url}
\usepackage{verbatim}
\usepackage{graphicx}
\usepackage[utf8]{inputenc}
\usepackage{cite}
\usepackage{microtype}
\usepackage{booktabs}
\usepackage{hyperref}
\usepackage{multirow}
\usepackage{float}
\usepackage{makecell}
\usepackage{algorithm}
\usepackage{algorithmic}
\begin{document}

\title{SPAGS: Sparse-View Articulated Object Reconstruction From Single State via Planar Gaussian Splatting}

\author{Di Wu, Liu Liu, Xueyu Yuan, Wenxiao Chen, Lijun Yue, Liuzhu Chen, Yiming Tang, Meng Wang~\IEEEmembership{Fellow,~IEEE} 
\thanks{Di Wu, Liu Liu, Xueyu Yuan, Wenxiao Chen, Yiming Tang and Meng Wang are with Hefei University of Technology  (E-mails: 18098697517@163.com; liuliu@hfut.edu.cn; yxyxc0410@mail.hfut.edu.cn; cassidy@mail.hfut.edu.cn; tym608@163.com; wangmeng@hfut.edu.cn).}
\thanks{Lijun Yue, Liuzhu Chen  are with Tencent RoboticsX (E-mails:  tabyue@tencent.com;nexchen@tencent.com ).}
\thanks{Corresponding author: Liu Liu.}
\thanks{This work has been submitted to the IEEE for possible publication. Copyright may be transferred without notice, after which this version may no longer be accessible.}
}



\maketitle

\begin{abstract}
Articulated objects are ubiquitous in daily environments, and their 3D reconstruction holds great significance across various fields. However, existing articulated object reconstruction methods typically require costly inputs such as multi-stage and multi-view observations.  To address the limitations, we propose a category-agnostic articulated object reconstruction framework via planar Gaussian Splatting, which only uses sparse-view RGB images from a single state. Specifically, we first introduce a Gaussian information field to perceive the optimal sparse viewpoints from candidate camera poses. To ensure precise geometric fidelity, we constrain traditional 3D Gaussians into planar primitives, facilitating accurate normal and depth estimation. The planar Gaussians are then optimized in a coarse-to-fine manner, regularized by depth smoothness and few-shot diffusion priors. Furthermore, we leverage a Vision-Language Model (VLM) via visual prompting to achieve open-vocabulary part segmentation and joint parameter estimation. Extensive experiments on both synthetic and real-world datasets demonstrate that our approach significantly outperforms existing baselines, achieving superior part-level surface reconstruction fidelity. Code and data are provided in the supplementary material.
\end{abstract}

\begin{IEEEkeywords}
3D Reconstruction, Computer Vision, Computer Graphics, Articulated Object
\end{IEEEkeywords}

\section{Introduction}
\label{sec:intro}

Articulated objects are prevalent in our daily life, such as drawers and scissors. Reconstructing articulated objects holds significant value across various fields, including embodied intelligence, virtual reality, and robotics. However, high-fidelity reconstruction of articulated objects is a non-trivial task since they vary greatly in size and require part-level mesh extraction for manipulation.

\begin{figure}[h]
    \centering
    \vspace{-8pt}
    \includegraphics[width=\linewidth]{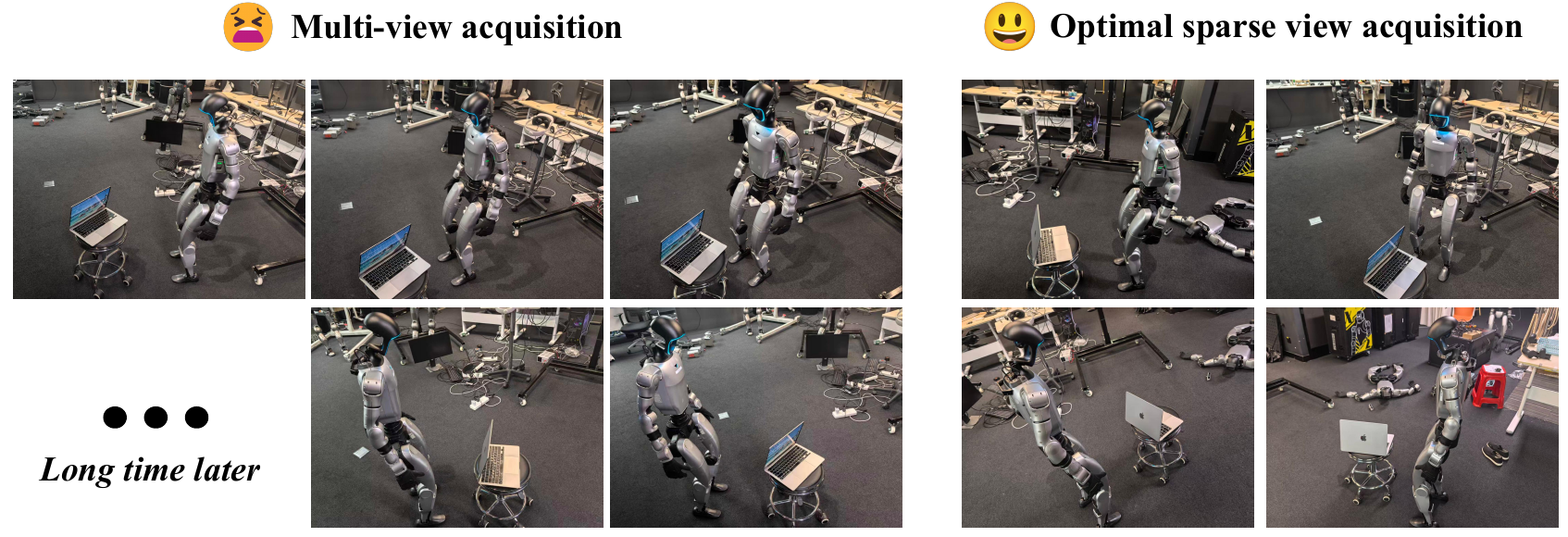}

    \caption{Comparison of the optimal sparse view acquisition and multi-view acquisition in robotic active reconstruction. Our method significantly reduces the complexity of the robot's movement and the inverse kinematics computation.}
    \vspace{-10pt}
    \label{fig:active_recon}
\end{figure}

Under this circumstance, some methods~\cite{mu2021learning, jiang2022ditto}  learn category-specific prior knowledge for articulated object reconstruction. Nevertheless, these models struggle with unseen object types.  Recently, PARIS~\cite{jiayi2023paris}, REArtGS~\cite{wu2025reartgs}, and ArtGS~\cite{liu2025building} achieve category-agnostic part-level reconstruction using multi-view RGB or RGBD images from two states of articulated objects. However, obtaining multi-view observations from two states is costly in practical applications, especially in the task of active 3D reconstruction for robots, which relies on time-consuming robot movements and complex inverse kinematics computation to complete image acquisition, which is illustrated in Fig.~\ref{fig:active_recon}.


\begin{figure}[h]
    \centering
    \includegraphics[width=\linewidth]{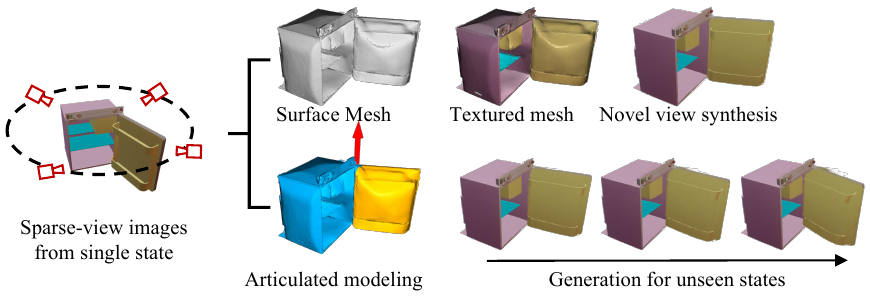}

    \caption{Given an arbitrary articulated object, our method enables autonomous optimal sparse viewpoints perception and produces: (1) surface mesh; (2) textured mesh; (3) novel view synthesis; (4) articulated modeling; (5) unseen state generation.}
    \label{fig:teaser}
    \vspace{-0.5em}
\end{figure}

In recent years, some approaches~\cite{zhu2023FSGS, paliwal2024coherentgs} are dedicated to reducing reliance on multi-view reconstruction. Unfortunately, when the input views become extremely sparse (e.g., 4 views), their reconstruction quality suffers a sharp decline. Most recently, GaussianObject~\cite{yang2024gaussianobject} incorporates diffusion model~\cite{song2020denoising} with 3D Gaussian Splatting (3DGS)~\cite{kerbl3Dgaussians}, achieving realistic object reconstruction only with 4-view RGB images. Nevertheless, we make observation that GaussianObject still exhibits followings limitations for articulated object reconstruction: (1) The disorderly and irregular nature of 3D Gaussian primitives makes depth and normal estimation difficult, leading to inaccurate surface reconstruction. (2) It requires a predefined view setting, and there is a deviation between the manually selected viewpoints and the optimal observation, affecting its application in robot autonomous perception. (3) It fails to extract part-level meshes, limiting downstream manipulation of articulated objects. 

To tackle the above-mentioned challenges, we propose SPAGS (shown in Fig.~\ref{fig:teaser}), the first category-agnostic framework of single-state \textbf{SP}arse-view \textbf{A}rticulated object reconstruction via \textbf{G}aussian \textbf{S}platting, to our best knowledge. Specifically, we first propose an optimal view perception method by establishing a Gaussian information field, which continuously estimates optimal viewpoints with the maximum information potential. We later acquire a structured initialization for Gaussian primitives via a 3D generative model~\cite{huang2025spar3d} and register the initialization using a pyramid network.

Afterwards, we compress 3D Gaussians into planar Gaussians to facilitate depth and normal estimation. We adopt a coarse-to-fine strategy to optimize the planar Gaussians.  In coarse training, we leverage the pseudo depth labels from a depth estimation model and depth smoothness regularization to enhance geometric learning. During refinement, we fine-tune a pretrained diffusion model using few-shot data by constructing image pairs within reliable regions, and then use the fine-tuned model to refine noisy regions. Subsequently, we employ a Visual Language Model (VLM) to generate an articulation tree, and assign  part-aware probability to each Gaussian primitive, which is updated via back-projecting 2D part masks. Then we perform joint parameter estimation via visual question answering. Extensive experiments demonstrate our method significantly outperforms the existing state-of-the-art (SOTA) methods on both synthetic and real-world data.


In summary, our main contributions can be summarized as follows:
\begin{itemize}
    \item We propose the first category-agnostic framework of high-fidelity articulated object reconstruction, using only sparse-view RGB images from a single state.
    \item We propose the Gaussian information field to estimate the information potential of a sampled camera pose, achieving optimal view perception.
    \item We propose a coarse-to-fine optimization strategy for planar Gaussians, improving geometry learning through depth smooth regularization and few-shot diffusion.  
\end{itemize}


\begin{figure*}[tbh]
    \centering
    \includegraphics[width=0.97\linewidth]{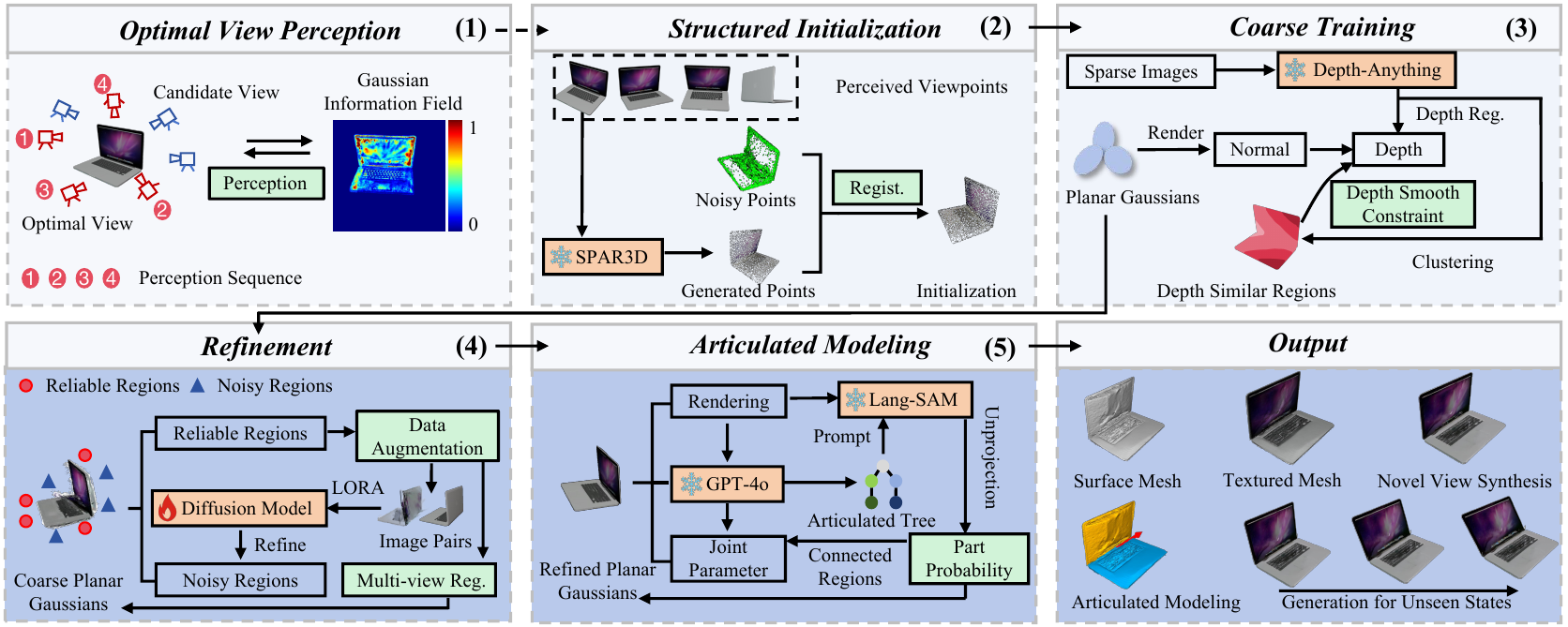}

    \caption{The Framework of SPAGS. We use the snowflake symbol to denote frozen network weights and the flame symbol to indicate trainable weights. ``Reg." and ``Regist." denote regularization and registration respectively. We highlight our main contributions in green. Note that the optimal view perception is an optional component. }
    \label{fig:method}
\end{figure*}

\section{Related Works}
\label{sec:formatting}

\subsection{Sparse-View Image Reconstruction}
While  3DGS~\cite{kerbl3Dgaussians} and its subsequent works~\cite{TMM1, TMM2, TMM3, TMM4, TMM5, TMM6, TMM7} emerge as novel fashions for high-fidelity scene reconstruction, these methods typically struggle with sparse-view input setting. Therefore, some works attempt to reduce the number of training views. Specifically,  RegNeRF~\cite{Niemeyer2021Regnerf} proposes additional geometric regularization from unobserved viewpoints. SparseNeRF~\cite{wang2022sparsenerf} distills local depth ranking priors from a monocular depth estimation model. Although achieving improved novel view synthesis results, they yield entangled representation from the implicit radiation fields, making it challenging to extract an acceptable surface mesh. Recently, FSGS~\cite{zhu2023FSGS} introduces a proximity-guided Gaussian unpooling to increase the density of 3D Gaussians with sparse-view training images. CoherentGS~\cite{paliwal2024coherentgs} proposes a coherent regularization to optimize structured 3D Gaussians. Sparse2DGS~\cite{Wu_2025_CVPR} combines MVS with 3DGS for improved sparse-view reconstruction. SparseGS~\cite{xiong2023sparsegs} and GaussianObject~\cite{yang2024gaussianobject} integrate the diffusion model with 3DGS for sparse 360-degree reconstruction. However, these methods still rely on pre-defined sparse-view selection, and the irregular nature of 3D Gaussian primitives hinders accurate normal and depth estimation. Recently, GauSS-MI~\cite{xie2025gaussmi}  leverages RGB observations for iterative next-best-view estimation. However, this  strategy requires image capture at every candidate pose, leading to a bottleneck in scenarios where image acquisition is infeasible. Therefore, achieving optimal viewpoint perception based solely on candidate camera poses offers significantly higher practical utility for real-world applications.

\subsection{Articulated Object Surface Reconstruction}
Articulated object surface reconstruction has gradually attracted more attention in recent years. ASDF~\cite{mu2021learning} and Ditto~\cite{jiang2022ditto}  learn pre-trained models for shape generation of articulated objects, but they struggle with unseen object types. SINGAPO~\cite{liu2024singapo} and DreamArt~\cite{lu2025dreamart} propose single-view surface generation methods for articulated objects. However, these methods usually yield unfaithful results in unseen regions. Recently, PARIS~\cite{jiayi2023paris}, ArticulatedGS~\cite{guo2025articulatedgs}, REArtGS~\cite{wu2025reartgs}, REArtGS++~\cite{wu2025reartgs++} and ArtGS~\cite{liu2025building} achieves category-agnostic high-fidelity reconstruction of articulated objects with two-stage multi-view RGB images. However, these methods still require costly inputs, limiting their practicality in downstream tasks. 

\section{Methodology}

As shown in Fig.~\ref{fig:method}, given an unseen articulated object with $q$ parts ($q\ge 2$), our method first perceives optimal sparse viewpoints from candidate camera poses that are randomly sampled, and then achieve part-level surface reconstruction.  We provide a detailed elaboration of our pipeline below.

\subsection{Optimal Viewpoint Perception and Representation Initialization}

As illustrated in Algorithm~\ref{alg:view-selection}, we randomly sample candidate camera poses $\mathcal{V} = \{\sigma_1, \dots, \sigma_N\}$
from the upper hemisphere of an arbitrary articulated object, and only select $K$ sparse views ($K \ll N$)  that maximize visual information gain for sparse 3DGS reconstruction.  The initial viewpoint set $\pi$ and 3D Gaussian primitives are initialized with a random viewpoint and corresponding captured image respectively. Generally, for the viewpoint set $\pi= \left \{1, ..., k \right \}$,  we introduce a reliability parameter $P$ for each 3D Gaussian primitive $\mathcal{G}_{i}$, and define the Gaussian Information Field (GIF) as following:
\begin{equation}
\Psi_{i} = -\log P(1:k) \cdot \alpha_{i} \prod_{j=1}^{i-1}\left(1-\alpha_{j}\right),
\label{eq:potential}
\end{equation}
where $\alpha$ is the $\alpha$-blending weight and $P(1:k) \in \left[0,1 \right]$ is current reliability, quantifying rendering quality via NIQE~\cite{NIQE} \textbf{without ground truth image}. We define the information potential $E$ as: $E = -\log P(1:k)$. Intuitively,  $\Psi$ encodes Gaussian reliable probability, forming a ray-modulated energy field.

For a candidate view $\sigma$, we define its Information Field Intensity (IFI) as $\mathcal{I}(\sigma)$, and is computed via path integration along camera rays $r$:
\begin{equation}
\mathcal{I}(\sigma) = \oint_{r(\sigma)} \! \Psi  \, \mathrm{d}r = \sum_{j=1}^{n_{\sigma}}\sum_{i=1}^{N_{\mathcal{G}}} \Psi_{i},
\label{eq:ifi}
\end{equation}
where $n$ means all measurement beams, and $N_{\mathcal{G}}$ denotes the number of the ordered Gaussians along the ray. We incorporate the $k+1$ optimal viewpoint $\sigma^{*}$ into existing viewpoint set $\pi$, and utilize the captured images from updated $\pi$ to optimize the Gaussian primitives by vanilla 3DGS pipeline~\cite{kerbl3Dgaussians}. Besides, we update $P$ through the accumulated information potential $E$ to ensure  $P$ within the defined bounds $[0,1]$. The accumulated information potential $E (1:k+1)$ is formulated as:
\begin{equation}
E(1:k+1) = E(1:k) + E(k+1)
\end{equation}
and $P(1:k+1)$ is updated by: 
\begin{equation}
P(1:k+1) = \text{exp}\left(-E \left(1:k+1 \right) \right)
\end{equation}

Notably, \textbf{the optimal viewpoint perception module is optional}. It eliminates manual selection bias when viewpoint choice is feasible, yet our approach sustains high-quality 3D reconstruction results under the constraints of fixed and sparse viewpoints, as demonstrated in Table.~\ref{tab:ablation_view}.

\begin{algorithm}[tb]
\caption{Optimal View Perception Algorithm}
\label{alg:view-selection}
\begin{algorithmic}[1]
\STATE Initialize $\mathcal{G}, \pi$ with a random view
\WHILE{$k<K$}
    \WHILE{$\sigma \in \mathcal{V}_{\text{remain}}$}
        \STATE $\mathcal{I}(\sigma) \gets \mathcal{G}, \sigma$
    \ENDWHILE
    \STATE $\sigma^* \gets \text{argmax}_{\sigma}\left( \mathcal{I}
    (\sigma) \right)$
    \STATE $\pi \gets \pi \cup \{\sigma^*\}$
    \STATE Optimize $\mathcal{G}$ with $\pi$
    \STATE Update $P$, $\mathcal{V}_{\text{remain}}$
\ENDWHILE 
\end{algorithmic}
\end{algorithm}



Once we finish sparse view perception, we attempt to obtain an initial  initialization $\mathbf{p} \in \mathbb{R} ^{3}$ using a monocular 3D generation model SPAR3D~\cite{huang2025spar3d}. However, this is not a ready-to-use pipeline since the generated pose and scale are inaccurate, which seriously affects subsequent reconstruction. We observe that although the  points $\mathbf{\bar{p}} \in \mathbb{R} ^{3}$ of Gaussian primitives from view perception are noisy, the global pose and scale of $\mathbf{\bar{p}}$ are relatively accurate.  Therefore, we seek to distill a structured representation with reasonable pose and scale through registering $\mathbf{p}$ and $\mathbf{\bar{p}}$. 

 Due to the complex disparities in pose and scale between $\mathbf{p}$ and $\mathbf{\bar{p}}$, we assume that each individual point $\mathbf{p}_{i} \in \mathbf{p}$ undergoes a similarity transformation  to $\mathbf{\hat{p}}_{i}$. We employ an exponential mapping $\mathbf{T} = \left ( \mathbf{\omega}, \mathbf{t}, s  \right ) $ to parameterize the similarity transformation,
where $s$ is the scaling factor and $\mathbf{t}$ is the translation. $\mathbf{\omega} \in \mathbb{R} ^{3} $ is an axis-angle vector. 
Inspired by~\cite{li2022DeformationPyramid}, we adopt a pyramid structure network to progressively learn the deformation parameters $\mathbf{T}$ which is decomposed into a sequence $ \left \{\mathbf{T}_{1}, \mathbf{T}_{2}, ..., \mathbf{T}_{l} \right \}$, with $l=9$. 
Please refer to the supplementary material for detailed elaboration.

\subsection{Coarse Training for Planar 3D Gaussians}
After obtaining the structured initialization $\mathbf{p}$ for Gaussian primitives, we first compress the 3D Gaissians to planar Gaussians, facilitating more reasonable depth and normal estimation. Concretely, we introduce a scale loss $\mathcal{L}_{\text{scale}}$ to flatten 3D Gaussian primitives into 2D planes.
\begin{equation}
\mathcal{L}_{\text{scale}} = \frac{1}{N_{\mathcal{G}}} \sum_{i}^{N_{\mathcal{G}}} \left \| \text{min}(S_{1}, S_{2}, S_{3}) \right \| 
\label{eq:scale}
\end{equation}
where $S_{1}, S_{2}, S_{3}$ denote the scale of Gaussian primitive $\mathcal{G}_{i}$ along each direction. Based on the planar Gaussian primitives, we can easily acquire the normal via the shortest axis and viewing direction. Denoting the normal rendering from $\alpha$-blending as $N$, we can derive the unbiased depth $\mathbf{D}$ following PGSR~\cite{chen2024pgsr}, which are illustrated in the supplementary materials.


To enhance geometric constraints for the planar Gaussians, we employ a monocular depth estimation model~\cite{depthanything} to introduce additional depth cues $\mathcal{D} $  for sparse-view images. Considering that the depth estimation result  $\mathcal{D}$ is typically noisy, we employ scale parameter $\boldsymbol{\varphi} = \left \{ \varphi_{1}, \varphi_{2}, ..., \varphi_{K} \right \}$ and offset parameter $\boldsymbol{\eta} = \left \{\eta_{1}, \eta_{2}, ..., \eta_{K} \right \}$ to adaptively refine the estimated depth  $\mathcal{D}_{i}$ for each view. Specifically, we set  $\boldsymbol{\varphi}$ as a learnable parameter to align depth scale. We optimize the parameters of an image convolutional decoder to update the offset $\boldsymbol{\eta}$ of whole image instead of estimating pixel-wise offset, ensuring a coherent variation. Given the depth rendering $\mathbf{D}$ of planar Gaussians, the depth regularization $\mathcal{L}_{\text{depth}}$ can be formulated as:
\begin{equation}
\mathcal{L}_{\text{depth}} = \sum_{i}^{K} \left \| \mathbf{D}_{i} - \left ( a_{i}\mathcal{D}_{i}+ \boldsymbol{\eta}_{i}  \right )   \right \|_{1} 
\label{eq:depth}
\end{equation}

Moreover, we encourage the depth variation of Gaussian primitives to be smooth. We start with adopting the $c$-channel segmentation module from~\cite{paliwal2024coherentgs} to segment each estimated depth $\mathcal{D}_{i}$ into $c$ similar regions, as shown in part (3) of Fig.~\ref{fig:method}. For each region, we introduce a local depth constraint  $\mathcal{L}_{\text{smooth}}$ to ensure the depth variations are smooth, which are expressed as:
\begin{equation}
\mathcal{L}_{\text{smooth}} = \sum_{i}^{K} \mathcal{E}(x_{i})\odot \sum_{j}^{c} \left (    \left | \nabla_{x}\mathbf{D}_{i}^{j}  \right | +\left | \nabla_{y}\mathbf{D}_{i}^{j}  \right |\right )
\label{eq:ds}
\end{equation}
where $\mathcal{E}$ is an edge detection model that returns binary edge mask for image $x_{i}$, using Sobel operator to eliminate the impact of depth smooth regularization on image edges.

The coarse training objective can be formulated as: $\mathcal{L}_{\text{coarse}} = \mathcal{L}_{\text{color}} + \mathcal{L}_{\text{planar}}$
, where $\mathcal{L}_{\text{color}}$ is the color loss of pixel space using in~\cite{kerbl3Dgaussians}. $\mathcal{L}_{\text{planar}}$ is expressed as:
\begin{equation}
\mathcal{L}_{\text{planar}} =  \lambda_{\text{scale}}\mathcal{L}_{\text{scale}}+\lambda_{\text{depth}}\mathcal{L}_{\text{depth}}+\lambda_{\text{smooth}}\mathcal{L}_{\text{smooth}}
\end{equation}
\subsection{Planar Gaussian Refinement via Few-shot Diffusion}
To optimize the representation of regions lacking supervision, we introduce the diffusion model~\cite{rombach2022high} to futher refine Gaussian primitives. Note that our approach only utilize few-shot data to fine-tune the diffusion model, i.e., the sparse-view inputs.  

Concretely, we first establish reliable regions $\Omega$ and noisy regions $\Omega^{\text{noisy}}$. For each  available viewpoint $\sigma_{i}$, we define the reliable regions $\Omega_{i}$ as:
\begin{equation}
\Omega_{i} = \left \{ \mathbf{R}_{i}^{\text{gen}}, \mathbf{t}_{i}^{\text{gen}} | \Delta \theta_{z},\Delta \theta_{y},\Delta \theta_{x}, \Delta \mathbf{t}] \right \}
\label{eq:reliable_regions}
\end{equation}
where $\mathbf{R}_{i}^{\text{gen}}$ and $\mathbf{t}_{i}^{\text{gen}}$ represent reliable generated rotations with a limited range for Euler angle perturbations $\Delta \theta_{z}$, $\Delta \theta_{y}$, $\Delta \theta_{x}$, and translations with perturbation $\Delta \mathbf{t}$ respectively. In turn, we define the camera poses that fall outside this range as the noisy regions. We then randomly sample a generated camera pose $\bar{\sigma}_{i}$ from each reliable region, and take its rendering $\bar{x}$  as pseudo label for the diffusion model.


\begin{figure}[h]
    \centering
    \includegraphics[width=\linewidth]{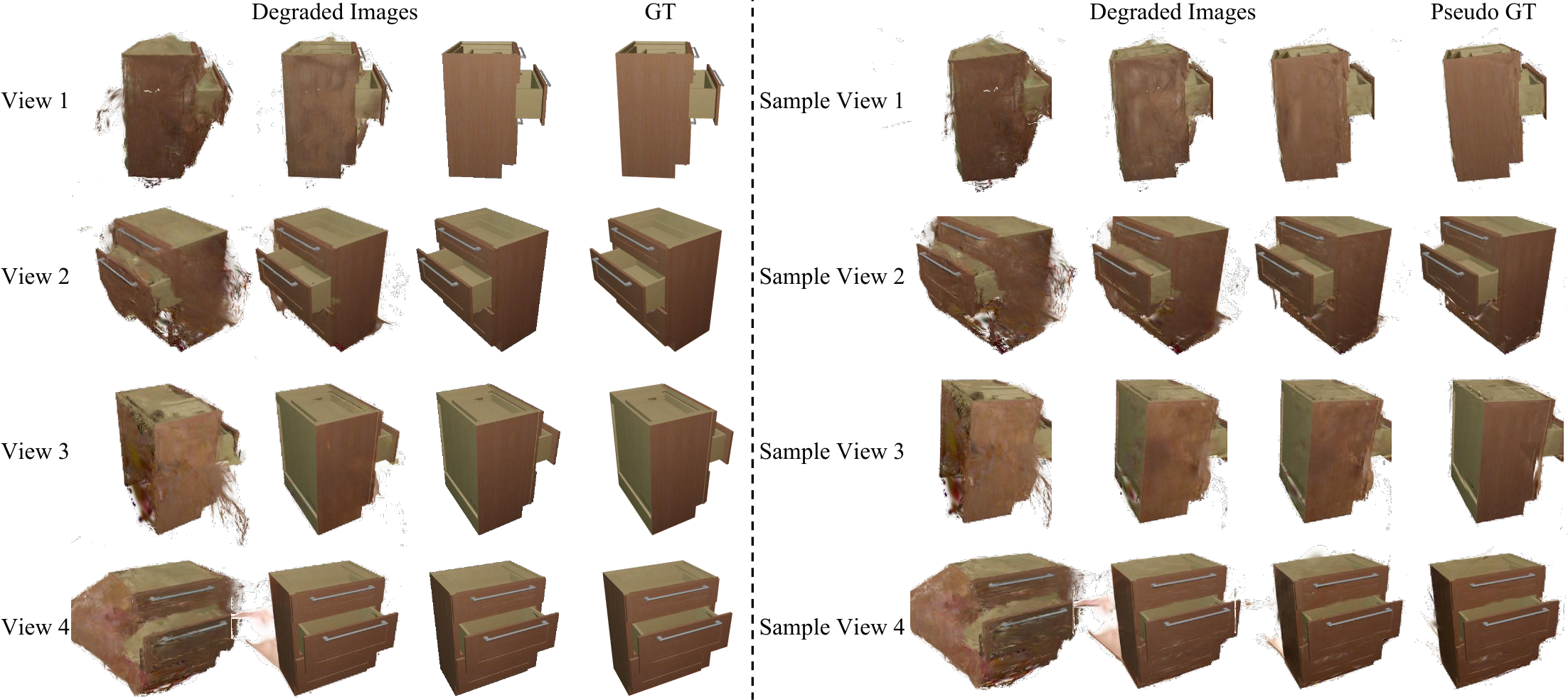}

    \caption{Illustration of data augmentation via few-shot image inputs.}
    \label{fig:data_aug}
\end{figure}

To augment the training data, we employ the leave-one-out strategy to generate additional image pairs, similar as~\cite{yang2024gaussianobject}. We divide the input data into $n$ subsets, each of which contains $n-1$ original images and one left-out image $x_{\text{left}}$. For each subset, we train the planar Gaussians for 6,000 iterations without $x_{\text{left}}$ to yield the degraded renderings $x_{\text{deg}}$ and $\bar{x}_{\text{deg}}$ for $x_{\text{left}}$ and its corresponding generated label $\bar{x}_{\text{left}}$ respectively. Subsequently, we incorporate the left-out image $x_{\text{left}}$ to continue training the Gaussian primitives 4,000 iterations, producing the repaired rendering for $x_{\text{deg}}$ and $\bar{x}_{\text{deg}}$. These renderings at different iterations are combined with $x_{\text{left}}$ and $\bar{x}_{\text{left}}$ to form image pairs for diffusion model fine-tuning, as shown in Fig.~\ref{fig:data_aug}.

We use a pre-trained ControlNet~\cite{zhang2023adding} to steer the diffusion model with conditional images. Specially, we add LoRA~\cite{hu2022lora} layers
to the transformer
blocks in the diffusion U-Net and the ControlNet U-Net. The fine-tuning objective $\mathcal{L}_{\text {diff }}$ can be formulated as:
\begin{equation}
\begin{aligned}
\mathcal{L}_{\text {diff }} &=\mathbb{E}\left[\left\|\left(\epsilon_{\theta}\left(x_{\text{left}}, t, x_{\text{deg}}\right)-\epsilon\right)\right\|_{2}^{2}\right]
\\&+\lambda_{\text{gen}} \mathbb{E}\left[\left\|\left(\epsilon_{\theta}\left(\bar{x}_{\text{left}}, t, \bar{x}_{\text{deg}}\right)-\epsilon\right)\right\|_{2}^{2}\right]
\label{eq:diffuson_train}
\end{aligned}
\end{equation}
where $\epsilon$ and $\epsilon_{\theta}$ represent the added noise and predicted noise respectively, and $t$ is the diffusion step.

Once the diffusion model fine-tuning is completed, we utilize it to optimize the noise regions $\Omega_{\text{noisy}}$. We start with randomly sampling viewpoints 
$\sigma_{\text{rand}}$ from $\Omega_{\text{noisy}}$, and obtain corresponding rendering $x(\mathcal{G}, \sigma_{\text{rand}})$. Following~\cite{meng2022sdedit}, we use a latent diffusion encoder $\boldsymbol{\varepsilon}$ to encode the rendering, later employ the image condition to yield a noisy latent representation $\mathbf{z}_{t}$:
\begin{equation}
\label{eq:diffusion}
\mathbf{z}_{t}=\sqrt{\bar{\alpha}_{t}} \boldsymbol{\varepsilon}\left(x(\mathcal{G}, \sigma_{\text{rand}})\right)+\sqrt{1-\bar{\alpha}_{t}} \epsilon
\end{equation}
where $\bar{\alpha}$ is a predefined coefficient. Then a denoised sample $x_{\text{gen}}$ can be derived from $\mathbf{z}_{t}$ through the diffusion model. Subsequently, we leverage the image pairs $(x_{\text{gen}}, x(\mathcal{G}, \sigma_{\text{rand}}))$ to refine the coarse Gaussian primitives $\mathcal{G}$, and the optimization loss $\mathcal{L}_{\text{repair}}$ is expressed as:
\begin{equation}
\mathcal{L}_{\text{repair}} = \left \| x_{\text{gen}} -  x(\mathcal{G}, \sigma_{\text{rand}}))   \right \|_{2} + f(x_{\text{gen}} -  x(\mathcal{G}, \sigma_{\text{rand}}) )
\label{eq:refine}
\end{equation}
where $f$ denotes the perceptual similarity function LPIPS.

Besides, we introduce a view consistency regularization $\mathcal{L}_{\text {vc}} $ to ensure the corresponding pixels rendered with adjacent viewpoints  originate from the same 3D points. Specifically, the 3D positions back-projected from the rendering pixel batch $\delta _{i} $ under viewpoint $\sigma_{i}$ should align with the corresponding rendering pixel batch under the generated camera pose $\bar{\sigma}_{i}$. Therefore, $\mathcal{L}_{\text {vc}} $ can be represented as following:
\begin{equation}
\mathcal{L}_{\text{vc}} = \frac{1}{ |\delta|   } \sum_{i} \left \|   \left(\Theta(\delta _{i}| \sigma_{i} )\right) - \left (\Theta (H\delta _{i}|\bar{\sigma}_{i})\right )  \right \| 
\label{eq:vc}
\end{equation}
where $\Theta $ denotes the back-projection operation via the depth rendering, and $H$ is the homography matrix follwing~\cite{chen2024pgsr}, mapping $\delta_{i}$ to its corresponding pixel batch under $\bar{\sigma}_{i}$. In this way, we enhance the geometric consistency of the planar Gaussians. The total training objective of the refinement is formulated as:
$\mathcal{L}_{\text{refine}} = \mathcal{L}_{\text{repair}} + \lambda_{\text{vc}}\mathcal{L}_{\text{vc}} +\mathcal{L}_{\text{planar}}$.

\begin{table*}[ht]
  \centering
  \caption{Exemplary Visual Question Answering (VQA) Sessions with the VLM.}
  \label{tab:SPAGS_vlm_prompts}
\resizebox{\linewidth}{!}{
  \begin{tabular}{m{4cm} | m{10.5cm}}
    \toprule
    \textbf{Task Objective} & \multicolumn{1}{c}{\textbf{VQA Interaction Examples}} \\
    \midrule

    Identify the topology of \newline  revolute joint & 
    \textbf{Prompt:} Analyze the rendered image. Does the physical hinge connect two distinct components (e.g., a container and its lid), or is it a surface-mounted pivot (e.g., a knob)? \newline 
    \textbf{Response:} The physical hinge establishes a connection between two distinct components. It is classified as type (1). \\
    \hline

    Revolute axis \newline candidate selection \newline (type 1) & 
    \textbf{Prompt:} Observe the rendered grid image with numbered spatial anchors. Select $n_{2} \geq 2$ points from the clustered articulation region to define the primary rotation axis. \newline 
    \textbf{Response:} Points 5 and 8 are spatially aligned with the physical hinge. I designate points [5, 8] to parameterize the rotation axis. \\
    \hline

    Revolute axis \newline candidate selection \newline (type 2) & 
    \textbf{Prompt:} Based on the identified articulation region and its estimated planar normal vector, select the optimal anchor point to determine the pivot's vertical axis. \newline 
    \textbf{Response:} The geometric center of the rotating knob serves as the optimal pivot. I select point [12]. \\
    \hline

    Identify the topology of \newline prismatic joint & 
    \textbf{Prompt:} Analyze the rendering. Does this mechanism represent an orthogonal sliding joint (e.g., a drawer) or a tangential surface sliding joint (e.g., a sliding blade)? \newline 
    \textbf{Response:} The component exhibits inward/outward translation relative to the base. It is identified as an orthogonal prismatic joint. \\
    \hline

    Surface sliding \newline direction alignment & 
    \textbf{Prompt:} Evaluate the directional arrows superimposed on the rendered image. Identify the index of the arrow that accurately represents the physical sliding trajectory. \newline 
    \textbf{Response:} The horizontal vector is congruent with the physical sliding track. The selected orientation corresponds to arrow [3]. \\

    \bottomrule
  \end{tabular}
}
\end{table*}

\subsection{Articulation Modeling for Planar Gaussians}
\begin{figure}[h]
    \centering
    \includegraphics[width=\linewidth]{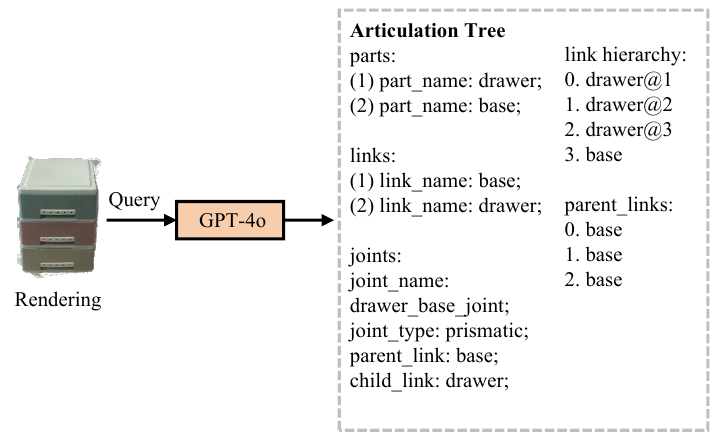}
    \caption{Illustration of joint estimation. Note that we use high-resolution rendering to query GPT-4o in actual inference.}
    \label{fig:art_tree}
\end{figure}

After the coarse-to-fine optimization, we employ GPT-4o to generate an abstract articulated tree for planar Gaussians, taking multiple rendered images as prompts, as shown in Fig.~\ref{fig:art_tree}. The articulated tree describes the name and connectivity of each part as well as the joint type.  




Subsequently, we introduce a part-aware probability $m_{i}$ for each Gaussian primitive $\mathcal{G}_{i}$. We render Gaussian primitives from a series of viewpoints to yield multiple rendered images, then we employ the Lang-SAM~\cite{lang_segment_anything} to generate 2D part-level  segmentation masks $\mathcal{M}$ for these renderings, using the part names from the articulated tree as text prompts. Following~\cite{chen2023gaussianeditor}, the probability $m_{i}^{o}$ of $\mathcal{G}_{i}$ belongs to part $o$ can be derived through back-projecting  $\mathcal{M}$ of corresponding pixel $\rho$:
\begin{equation}
m_{i}^{o}=\mathcal{M}^{o}(\rho )\alpha_{i} \prod_{j=1}^{i-1}\left(1-\alpha_{j}\right)
\end{equation}

In accordance with whether $m_{i}^{o}$ exceeds a predefined threshold $\tau $, we are able to determine whether the Gaussian belongs to part $o$. When finishing the part-level segmentation of the planar Gaussians, we employ TSDF Fusion to extract the surface mesh and set voxel size to 0.004.

\begin{table*}[h]
\centering
\small
\caption{Quantitative results for the surface reconstruction quality on PARIS dataset. $^{*}$ denotes methods designed for multi-view reconstruction.}
\resizebox{\linewidth}{!}{
\begin{tabular}{c|c|ccccccccccc}
\toprule

Metrics   &Method      & Stapler  & USB &  Scissor & Fridge & Foldchair & Washer & Blade & Laptop & Oven & Storage & Mean \\
\hline

\multirow{5}{*}{CD $\downarrow$}
& PGSR$^{*}$~\cite{chen2024pgsr} & 13.19  &26.94 &6.01 &672.76 &32.20 &32.40 &2.49 &24.12 &48.68 &81.90 &94.07\\

& CoherentGS~\cite{paliwal2024coherentgs} & 324.65  &33.76  & 313.04 &40.87 &80.61  &64.31 &9.34 &164.08 &41.42 &303.67 &137.58\\
&Sparse2DGS~\cite{Wu_2025_CVPR} &49.25 &35.57 &28.96 &38.87 &12.64 &425.27 &4.13 &28.47 &73.74 &60.88 &75.78\\

& GaussianObject~\cite{yang2024gaussianobject} & 7.39 &6.92 & \textbf{0.82} &36.05  &\textbf{0.58} &68.87 &1.58 &166.03 &\textbf{12.30} &126.93 &42.75\\
& SPAGS (Ours) &\textbf{5.65} & \textbf{6.56} &0.89 &\textbf{9.04} &0.66 &\textbf{27.54} &\textbf{1.05} &\textbf{5.21} &17.56 &\textbf{21.80} &\textbf{9.60}\\

\hline
\multirow{5}{*}{F1 $\uparrow$}
& PGSR$^{*}$~\cite{chen2024pgsr} & 0.08  &0.13 &0.33 &0.02 & 0.15 &\textbf{0.05} &0.25 &\textbf{0.19} &\textbf{0.03} &0.01 &0.12\\
& CoherentGS~\cite{paliwal2024coherentgs} &0.01  &0.06  &0.08 &\textbf{0.06} &0.15 &0.03 &\textbf{0.32} &0.17 &0.02 &0.00 &0.09 \\
&Sparse2DGS~\cite{Wu_2025_CVPR} &0.11 &0.07&0.24 &0.05 &0.18 &0.03 &0.16 &0.13 &\textbf{0.03} &0.04 &0.10\\
& GaussianObject~\cite{yang2024gaussianobject} &0.10 &0.15 &0.38 &0.05 &0.35 &\textbf{0.05} &0.27 &0.10 &\textbf{0.03} &0.01 &0.15 \\
& SPAGS (Ours) & \textbf{0.14} &\textbf{0.19} &\textbf{0.40} &\textbf{0.06} &\textbf{0.38} &\textbf{0.05} &\textbf{0.32} &0.13 &\textbf{0.03} &\textbf{0.02} &\textbf{0.17}\\

\hline
\multirow{5}{*}{EMD $\downarrow$}  
& PGSR$^{*}$~\cite{chen2024pgsr} & 0.14  &0.19 &\textbf{0.07} &0.84  &0.16 &\textbf{0.17} &0.06 &0.13 &0.22 &0.23 &0.22\\
& CoherentGS~\cite{paliwal2024coherentgs} & 0.53 &0.26 &0.44 &0.20 &0.32 &0.23 &0.06  &0.16 &0.25 &0.60  &0.31\\
&Sparse2DGS~\cite{Wu_2025_CVPR} &0.50 &0.59 &0.64 &0.62 &0.30 &0.72 &0.50 &0.51 &0.28 &0.31 &0.50\\
& GaussianObject~\cite{yang2024gaussianobject} &0.17 &0.12 &0.09 &0.12 &0.09 &0.20 &0.05 &0.31 &0.17 &0.38 &0.17\\
& SPAGS (Ours) & \textbf{0.12} &\textbf{0.11} &0.09 &\textbf{0.10} &\textbf{0.08} &\textbf{0.17} &\textbf{0.04} &\textbf{0.10} &\textbf{0.16} &\textbf{0.20} &\textbf{0.12}\\

\bottomrule
\end{tabular}
}

\label{tab:mesh_reconstruction}
\end{table*}

\begin{figure}[h]
    \centering
    \includegraphics[width=\linewidth]{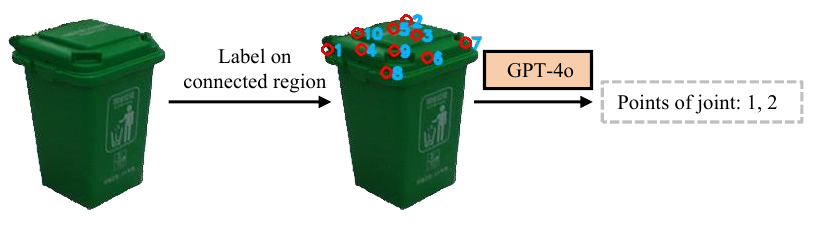}

    \caption{Illustration of joint estimation. Note that we use high-resolution rendering to query GPT-4o in actual inference.}
    \label{fig:joint_estimation}
\end{figure}

After obtaining the part-level Gaussian primitives, we utilize Visual Question Answering (VQA) to achieve open-vocabulary articulated parameter estimation, as shown in Table.~\ref{tab:SPAGS_vlm_prompts}. 
Following Articulated-Anymesh~\cite{qiu2025articulate}, we categorize revolute joints via a VLM (GPT-4o~\cite{openai2024gpt4o}) into two types based on their topology: 
(1) hinge-connected parts and (2) surface-anchored axes (e.g., knobs). 
For type (1), we cluster $n_1 \geq 10$ candidate points in the junction area, project labeled points onto 2D renderings, and prompt the VLM to select $n_2 \geq 2$ points to define the axis, as shown in Fig.~\ref{fig:joint_estimation}. The joint pivot is computed as the mean position, with the direction derived via Singular Value Decomposition (SVD) on the point offsets. For type (2), the joint is defined by the plane normal and a selected pivot point. 

Similarly, prismatic joints are functionally classified into inward/outward or surface-sliding types. The former's direction is determined by the fitted plane normal of the connection area, while the latter is identified through VLM-based selection among visual arrow prompts. Please refer to the supplementary materials for a detailed elaboration.

\begin{table*}[h]
\centering
\caption{Quantitative results for the novel view synthesis on PARIS dataset. $^{*}$ denotes methods designed for multi-view reconstruction.}
\resizebox{\linewidth}{!}{
\begin{tabular}{c|c|ccccccccccc}
\toprule

Metrics   &Method      & Stapler  & USB &  Scissor & Fridge & Foldchair & Washer & Blade & Laptop & Oven & Storage & Mean \\
\hline

\multirow{5}{*}{PSNR $\uparrow$}
& PGSR$^{*}$~\cite{chen2024pgsr} &19.68 & 19.55 &24.75 &11.90 &12.70 &17.60 &27.82 &22.69 &17.36 &17.85 &19.19\\
& CoherentGS~\cite{paliwal2024coherentgs} & 11.54 & 11.99 &13.60 &12.85 &9.31 &15.12 &16.88 &12.64 &13.63 &7.93 &12.55\\
&Sparse2DGS~\cite{Wu_2025_CVPR} &21.80 &17.51 &25.15 &22.00 &22.29 &19.79 &27.79 &16.07 &19.12 &22.21 &21.37\\
& GaussianObject~\cite{yang2024gaussianobject} &23.42 &22.17 &25.52 &23.11 &20.94  &19.04 &28.85 &21.46 
 &\textbf{20.53} &14.41 &21.95\\
& SPAGS (Ours) &\textbf{26.63} &\textbf{24.62} &\textbf{27.19} &\textbf{23.38} &\textbf{23.14} &\textbf{20.48} &\textbf{32.46} &\textbf{22.71} &20.07 &\textbf{20.63} &\textbf{24.13}\\

\hline
\multirow{5}{*}{SSIM $\uparrow$}
& PGSR$^{*}$~\cite{chen2024pgsr} & 0.93 &0.91 &0.95 &0.63 &0.76 &0.89 &0.97 &\textbf{0.94} &0.84 &0.86 &0.87\\
& CoherentGS~\cite{paliwal2024coherentgs} & 0.85 &0.83 &0.92 &0.82  &0.76 &0.86 &0.94 &0.91 &0.77 &0.57 &0.82\\
&Sparse2DGS~\cite{Wu_2025_CVPR} & 0.96 &0.90 &0.96 &0.83 &0.89 &0.70  &0.98 &0.91 &\textbf{0.89} &\textbf{0.91} &0.89\\
& GaussianObject~\cite{yang2024gaussianobject} &0.96 &0.92 &0.96 &0.92 &0.91 &0.88 &0.98 &0.91 &0.87 &0.78 &0.90 \\
& SPAGS (Ours) & \textbf{0.98} &\textbf{0.96} &\textbf{0.97} &\textbf{0.95} &\textbf{0.92} &\textbf{0.94} &\textbf{0.99} &\textbf{0.94} &\textbf{0.89} &0.89 &\textbf{0.94}\\

\hline
\multirow{5}{*}{LPIPS $\downarrow$}  
& PGSR$^{*}$~\cite{chen2024pgsr} & 0.07 &0.10 &0.06 &0.43 &0.22 &0.13 &0.02 &\textbf{0.07} &0.19 &0.22 &0.15\\
& CoherentGS~\cite{paliwal2024coherentgs} & 0.17 &0.19 &0.10 &0.26  &0.25 &0.20 &0.09 &0.09 &0.27  &0.53 &0.22\\
&Sparse2DGS~\cite{Wu_2025_CVPR} &0.06 &0.09 &0.12 &0.12 &0.10 &0.11 &0.03 &0.14 &0.12 &\textbf{0.15} &0.10\\
& GaussianObject~\cite{yang2024gaussianobject} &0.04 &0.12 &\textbf{0.03} &0.09 &\textbf{0.07} &0.15 &0.02 &0.09 &0.14 &0.29 &0.10 \\
& SPAGS (Ours) &\textbf{0.03} &\textbf{0.05} &\textbf{0.03} &\textbf{0.07} &\textbf{0.07} &\textbf{0.10} &\textbf{0.01} &\textbf{0.07} &\textbf{0.13} &0.18 &\textbf{0.07}\\

\bottomrule
\end{tabular}
}

\label{tab:nvs}
\end{table*}

\begin{table*}[h]
\centering
\small

\caption{Quantitative results for articulated modeling on the PARIS dataset.}
\label{tab:SPAGS_paris}
\resizebox{\linewidth}{!}{
\begin{tabular}{ccccccccccccc}
\toprule

\multirow{2}{*}{Metrics} & \multirow{2}{*}{Method} & \multirow{2}{*}{Stapler} & \multirow{2}{*}{USB} & \multirow{2}{*}{Scissors} & \multirow{2}{*}{Fridge} & \multirow{2}{*}{\shortstack{Foldchair}} & \multirow{2}{*}{\shortstack{Washer}} & \multirow{2}{*}{Oven} & \multirow{2}{*}{Laptop} & \multirow{2}{*}{\shortstack{Blade}} & \multirow{2}{*}{Storage} & \multirow{2}{*}{Mean} \\
& & & & & & & & & & & & \\

\midrule
\multirow{3}{*}{\shortstack{Axis\\Ang.}$\downarrow$}
& ArtGS~\cite{liu2025building} & 30.14 & 76.48 & 84.35 & 44.51 & 18.63 & 78.05 & 86.37 & 15.48 & 26.81 & 62.35 & 52.32 \\
& REArtGS++~\cite{wu2025reartgs++} & 40.24 & 75.27 & 62.15 & 66.34 & 83.17 & 68.24 & 75.78 & 42.04 & 61.50 & 36.02 & 61.07 \\
& SPAGS (Ours)& \textbf{4.01} & \textbf{0.07} & \textbf{0.03} & \textbf{0.02} & \textbf{0.25} & \textbf{5.13} & \textbf{7.48} & \textbf{0.04} & \textbf{3.11} & \textbf{0.14} & \textbf{2.03} \\

\midrule
\multirow{3}{*}{\shortstack{Axis\\Pos.}$\downarrow$}
& ArtGS~\cite{liu2025building} & 0.16 & 0.31 & 0.24 & 0.54 & 0.09 & 0.16 & 0.39 & 0.07 & - & - & 0.20 \\
& REArtGS++~\cite{wu2025reartgs++} & 0.19 & 0.25 & 0.27 & 0.59 & 0.11 & 0.24 & 0.43 & 0.05 & - & - & 0.21 \\
& SPAGS (Ours)& \textbf{0.09} & \textbf{0.01} & \textbf{0.05} & \textbf{0.02} & \textbf{0.03} & \textbf{0.11} & \textbf{0.02} & \textbf{0.01} & - & - & \textbf{0.03} \\

\midrule
\multirow{3}{*}{\shortstack{CD-s}$\downarrow$}
& ArtGS~\cite{liu2025building} & \textbf{62.65} & 64.01 & 77.51 & 99.61 & 195.07 & 271.24 & 100.34 & 192.04 & 16.27 & 25.10 & 110.38 \\
& REArtGS++~\cite{wu2025reartgs++} & 89.51 & 52.20 & 107.36 & 85.47 & 103.98 & 184.71 & 115.60 & 218.51 & 26.44 & 36.69 & 102.05 \\
& SPAGS (Ours)& 66.82 & \textbf{42.01} & \textbf{8.80} & \textbf{48.63} & \textbf{23.44} & \textbf{18.25} & \textbf{53.74} & \textbf{21.75} & \textbf{1.80} & \textbf{24.92} & \textbf{31.02} \\

\midrule
\multirow{3}{*}{\shortstack{CD-m}$\downarrow$}
& ArtGS~\cite{liu2025building} & 137.51 & 28.47 & 49.75 & 158.24 & 134.29 & 507.07 & 543.75 & 104.01 & 85.84 & 178.21 & 192.71 \\
& REArtGS++~\cite{wu2025reartgs++} & 98.57 & 56.28 & 32.41 & 128.09 & 204.30 & 473.19 & 489.21 & 63.37 & 131.76 & 171.98 & 184.92 \\
& SPAGS (Ours)& \textbf{27.74} & \textbf{6.00} & \textbf{6.26} & \textbf{23.94} & \textbf{0.95} & \textbf{5.69} & \textbf{95.14} & \textbf{8.44} & \textbf{17.85} & \textbf{66.72} & \textbf{25.87} \\

\midrule
\multirow{3}{*}{\shortstack{CD}$\downarrow$}
& ArtGS~\cite{liu2025building} & 85.12 & 58.60 & 23.45 & 120.30 & 45.70 & 201.10 & 185.00 & 110.20 & 48.90 & 130.40 & 100.90 \\
& REArtGS++~\cite{wu2025reartgs++} & 71.50 & 49.30 & 18.90 & 105.10 & 39.80 & 178.60 & 162.20 & 95.70 & 35.60 & 110.90 & 86.80 \\
& SPAGS (Ours) &\textbf{5.65} & \textbf{6.56} &0.89 &\textbf{9.04} &0.66 &\textbf{27.54} &17.56 &\textbf{5.21} &\textbf{1.05} &\textbf{21.80} &\textbf{9.60}\\

\bottomrule
\end{tabular}
}
\end{table*}

\section{Experiments}

\subsection{Experimental Setting}
\label{sec:exp_setting}
The comparisons include SOTA methods: PGSR~\cite{chen2024pgsr} (planar Gaussians), ArtGS~\cite{liu2025building}, REArtGS++~\cite{wu2025reartgs++} (category-agnostic reconstruction), CoherentGS~\cite{paliwal2024coherentgs}, Sparse2DGS~\cite{Wu_2025_CVPR}, GaussianObject~\cite{yang2024gaussianobject} (sparse-view reconstruction). Except for ArtGS and REArtGS++, all other methods are trained using the same 4-view RGB images as ours, while ArtGS and REArtGS++ are provided with additional 4-view images from another motion state. 

For whole object reconstruction, we use Chamfer Distance (CD), F1-score, Earth Mover's Distance (EMD)~\cite{zhang2022deepemd} as the metrics of mesh quality, and use PSNR, SSIM, LPIPS~\cite{zhang2018perceptual} as novel view synthesis metric. For articulated modeling, we employ CD-s and CD-m for surface meshes of static parts and dynamic parts respectively. We evaluate the predicted joints using angular error (Axis Ang.) and pivot distance (Axis Pos., for revolute joints only). Note that the CD resuls are multiplied by 1,000.

We conduct all experiments on a single RTX 4090 GPU. Please refer to the supplementary materials for detailed implementation and metrics calculation.


\textbf{Datasets}.
We conduct our experiments on both synthesis and real-world data. For synthesis data, we select the same 10 categories from  PARIS dataset~\cite{xiang2020sapien} and 5 categories from ArtGS-Multi dataset~\cite{liu2025building} as REArtGS++. Specifically, the PARIS dataset contains articulated objects composed of two parts, whereas the ArtGS-Multi dataset comprises multi-part articulated objects.  
Each object contains 64 randomly sampled candidate camera poses and corresponding RGB images.

For real-world data, we scan 6 categories of articulated objects, characterized by a diverse scale and geometry. Each object contains 76 to 160 randomly sampled candidate camera poses and corresponding RGB images captured in unbounded indoor scenes. We apply the optimal view selection algorithm to determine 4 optimal sparse views for each object in the two datasets, and use SAM~\cite{kirillov2023segany} to generate object masks in real-world data. Note that we only use the candidate camera poses without images in view perception.





\begin{figure}[h]
    \centering
    \includegraphics[width=\linewidth]{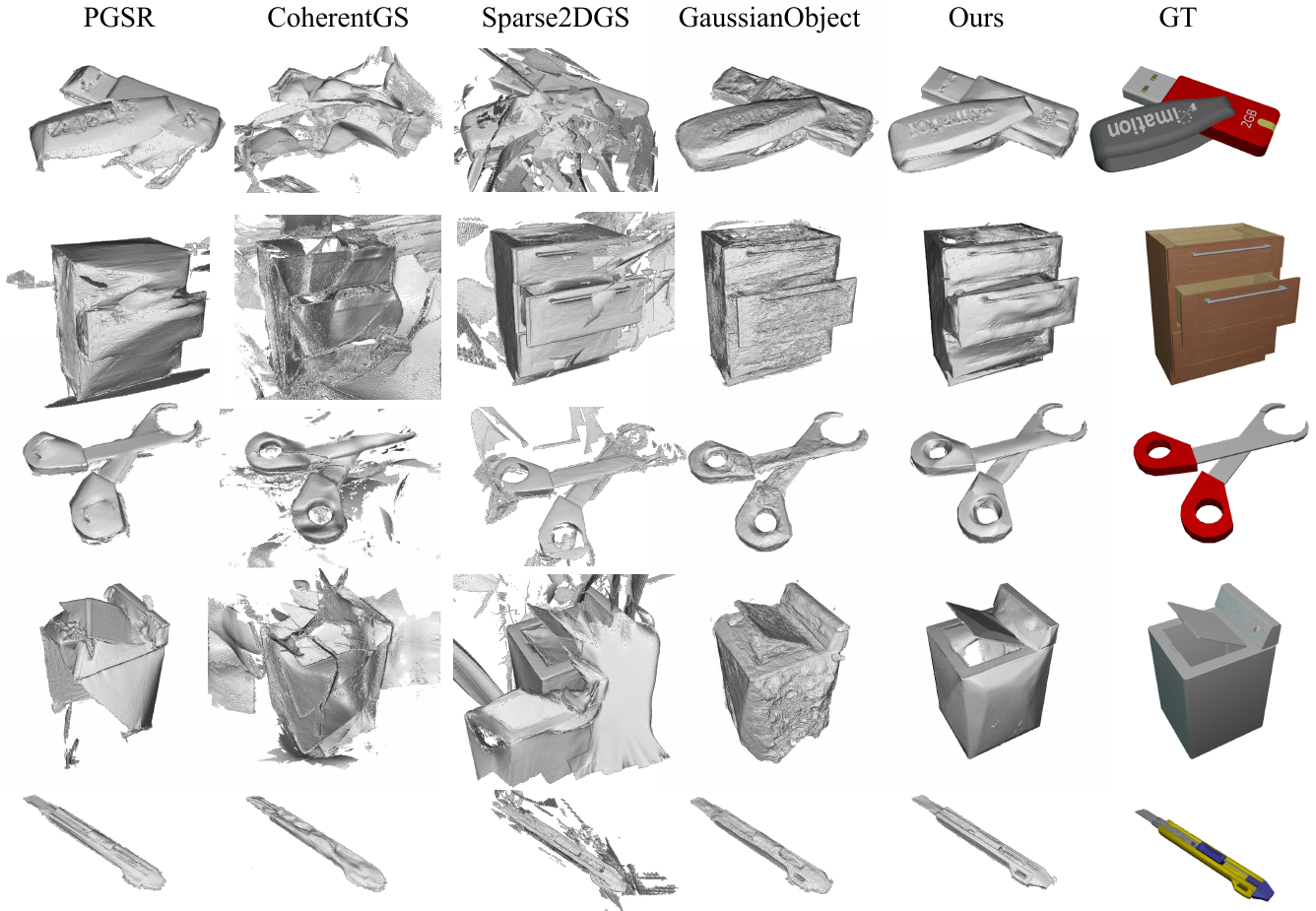}

    \caption{The qualitative results of whole mesh reconstruction on PARIS dataset.}

    \label{fig:mesh_rendeing1}
\end{figure}

\begin{figure}[h]
    \centering
    \includegraphics[width=\linewidth]{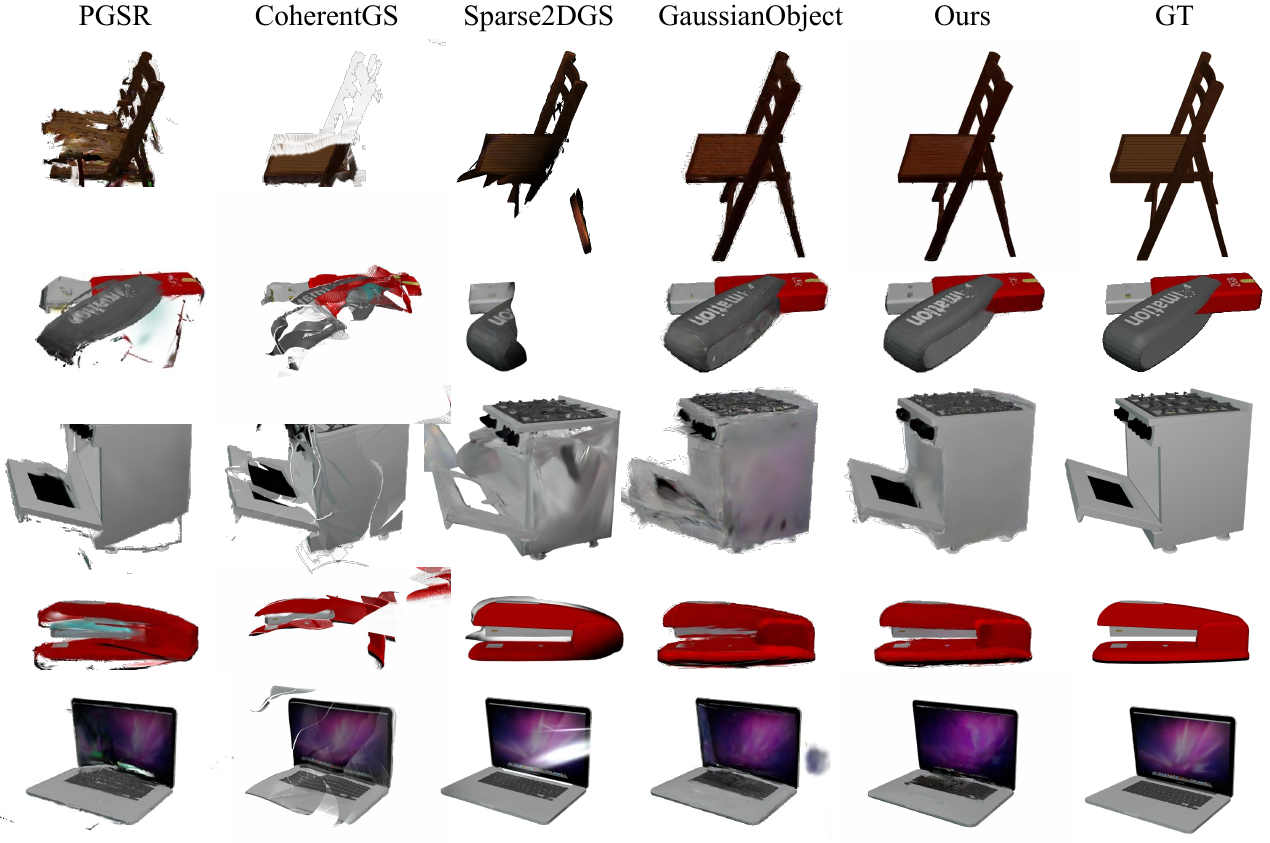}

    \caption{The  qualitative results of novel view synthesis on PARIS dataset.}
    \vspace{-1em}
    \label{fig:nvs1}
\end{figure}

\begin{figure}[h]
    \centering
    
    \includegraphics[width=0.95\linewidth]{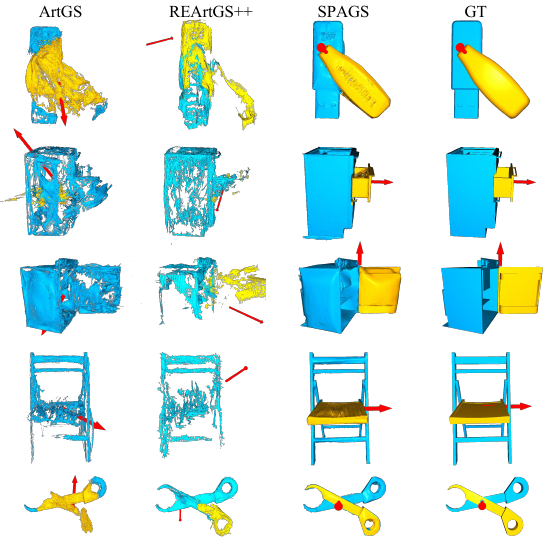}

    \caption{The qualitative results of articulated modeling on PARIS dataset.}
    \label{fig:seg}
\end{figure}

\subsection{Comparison with State-of-the-Art Methods}

\textbf{Mesh Reconstruction.}
We use the Chamfer distance (CD), F1-score, and EMD~\cite{zhang2022deepemd} as the metrics for surface mesh quality. We provide the quantitative results in Table.~\ref{tab:mesh_reconstruction} and present qualitative results in Fig.~\ref{fig:mesh_rendeing1}. More qualitative results can be found in the supplementary materials. Our method achieve bests mean results across all metrics with \textbf{9.60},  \textbf{0.17} and  \textbf{0.12},  outperforming other methods significantly. We observe that Sparse2DGS exhibits lots of artifacts under sparse view supervisions, as illustrated in Fig.~\ref{fig:mesh_rendeing1}. GaussianObject and CoherentGS produce surface meshes with lots of artifacts, primarily caused by the lack of geometric constraints. In contrast, our method maintains high-fidelity reconstruction by the coarse-to-fine optimization of planar Gaussians.

\begin{figure}[h]
    \centering
   
    \includegraphics[width=\linewidth]{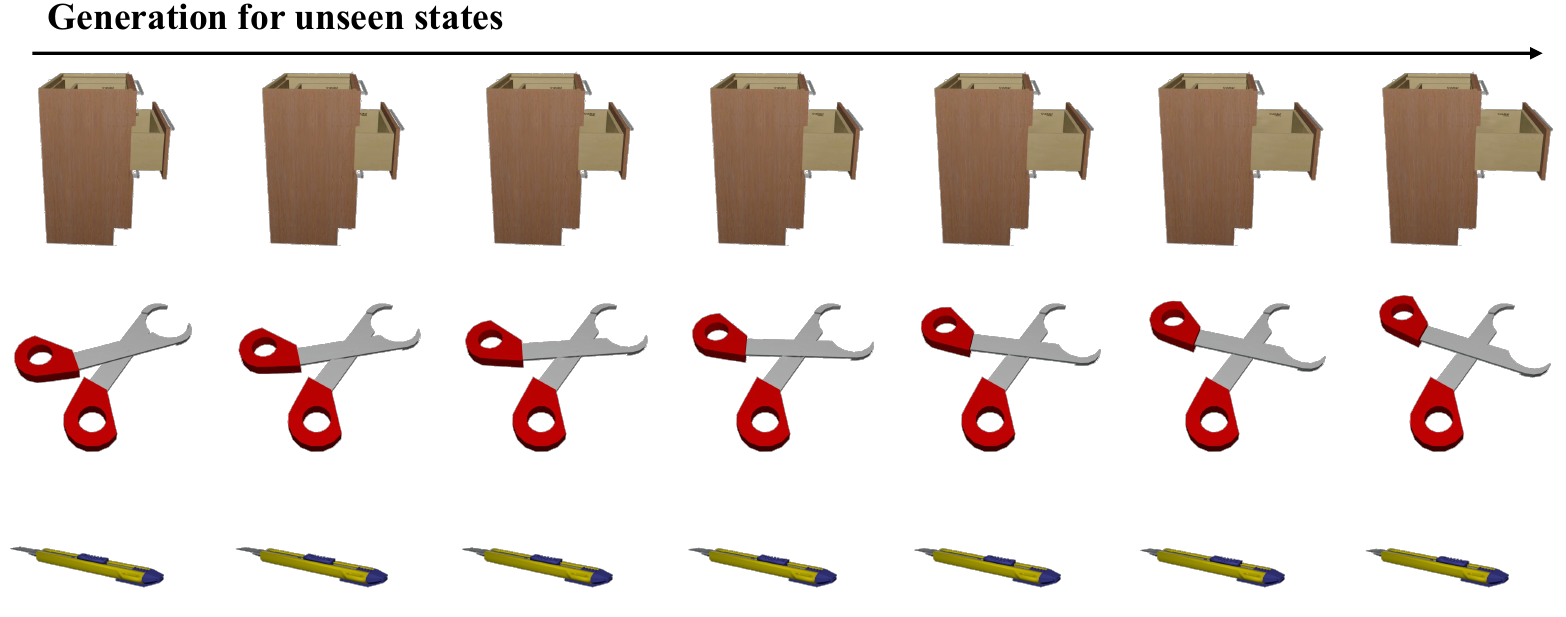}

    \caption{Additional qualitative results of generation at unseen states on PARIS dataset.}
    \label{fig:dynamic_render}
\end{figure}

\textbf{Novel View Synthesis Performance.}
We employ PSNR, SSIM, and LPIPS as metrics for novel view synthesis. Table.~\ref{tab:nvs} shows the quantitative results, and Fig.~\ref{fig:nvs1} presents qualitative comparisons. More qualitative results can be found in the supplementary materials. Our method achieves best results across all metrics with \textbf{24.13}, \textbf{0.94} and \textbf{0.07}. As shown in Fig.~\ref{fig:nvs1}, our method exhibits more realistic rendering results, while other methods produce significant artifacts and noise in renderings, especially for Sparse2DGS and CoherentGS. This is mainly because our method enhances noise region reconstruction through few-shot diffusion. Moreover, we provide the qualitative results of generation at unseen states, as shown in Fig.~\ref{fig:dynamic_render}.

\begin{figure}[h]
    \centering
    \includegraphics[width=\linewidth]{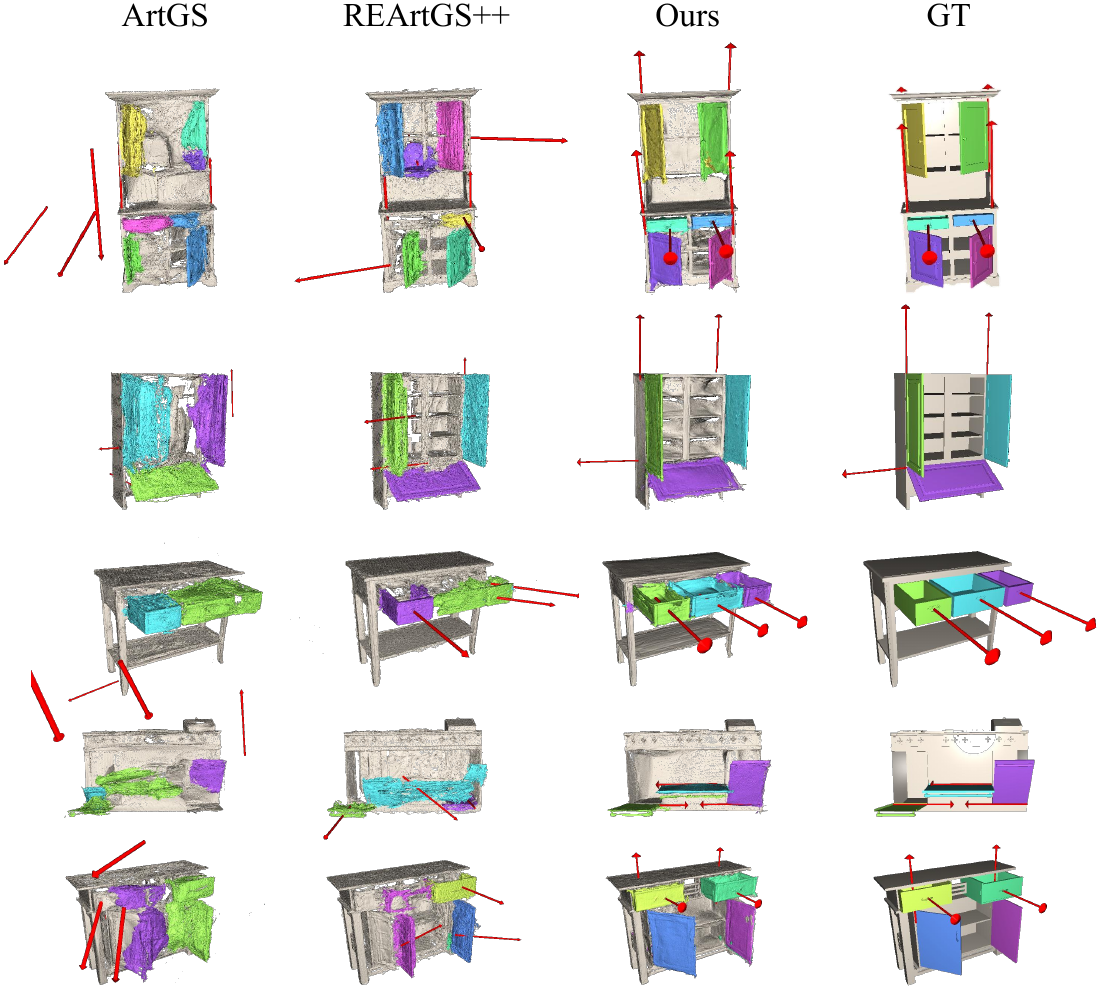}

    \caption{The qualitative results of articulated modeling on ArtGS-Multi dataset.}
    \label{fig:artgs_multi}

\end{figure}

\textbf{Articulated Modeling.}
On the basis of the overall CD, we additionally compare CD-m, CD-s, Axis Ang. and Axis Pos., as illustrated in Sec.~\ref{sec:exp_setting}. The quantitative results of articulated modeling on PARIS dataset are presented in  Table.~\ref{tab:SPAGS_paris} and the qualitative results are shown in Fig.~\ref{fig:seg}. \textbf{We also provide quantitative results of joint estimation in the supplementary materials.} Although ArtGS and REArtGS++ utilize two-stage observations, our method still achieve bests mean results on both CD-m and CD-s with \textbf{25.87},  \textbf{31.02}, significantly exceeding them. As shown in Fig.~\ref{fig:seg}, our method yields high-fidelity part-level mesh reconstruction and accurate joint estimation. This demonstrates our part segmentation and joint estimation for planar Gaussians are simple but effective, achieving realistic results without two-stage observations.

To further evaluate the articulated modeling performance for multi-part objects, we conduct experiments on the ArtGS-Multi dataset. The quantitative and qualitative results are presented in Table.~\ref{tab:SPAGS_artgs} and Fig.~\ref{fig:artgs_multi} respectively. Experimental results validate the robustness and superiority of our method for modeling multi-part articulated objects.

\begin{table}[h]
\centering
\small

\caption{Quantitative results for articulated modeling  on ArtGS-Multi dataset.}
\label{tab:SPAGS_artgs}

\resizebox{\linewidth}{!}{
\begin{tabular}{cccccccc}
\toprule
\multirow{2}{*}{Metrics} & \multirow{2}{*}{Method} 
& Storage & Storage & Table & Table & Oven & \multirow{2}{*}{Mean} \\
& &\scriptsize{47468 (7 parts)} &\scriptsize{45503 (4 parts)} &\scriptsize{25493 (4 parts)} &\scriptsize{31249 (5 parts)} &\scriptsize{101908 (4 parts)} & \\

\midrule
\multirow{3}{*}{\shortstack{Axis\\Ang.}$\downarrow$}
&ArtGS~\cite{liu2025building} &65.40 &82.10 &45.20 &38.70 &50.30 &56.34\\
&REArtGS++~\cite{wu2025reartgs++}  &38.20 &45.60 &35.40 &32.50 &28.90 &36.12\\
&SPAGS (Ours)&\textbf{4.20} &\textbf{6.80} &\textbf{3.50} &\textbf{4.10} &\textbf{2.90} &\textbf{4.30}\\
\midrule
\multirow{3}{*}{\shortstack{Axis\\Pos.}$\downarrow$}
&ArtGS~\cite{liu2025building} &2.05 &2.30 &- &1.15 &1.70 &1.80\\
&REArtGS++~\cite{wu2025reartgs++}  &1.55 &1.62 &- &1.38 &1.40 &1.49\\
&SPAGS (Ours)&\textbf{0.14} &\textbf{0.22} &- &\textbf{0.11} &\textbf{0.08} &\textbf{0.14}\\
\midrule
\multirow{3}{*}{\shortstack{CD-s}$\downarrow$}
&ArtGS~\cite{liu2025building} &80.10 &78.30 &62.50 &55.40 &68.20 &68.90\\
&REArtGS++~\cite{wu2025reartgs++}  &52.40 &48.50 &45.20 &42.60 &46.80 &47.10\\
&SPAGS (Ours)&\textbf{16.50} &\textbf{18.20} &\textbf{15.40} &\textbf{16.10} &\textbf{14.80} &\textbf{16.20}\\
\midrule
\multirow{3}{*}{\shortstack{CD-m}$\downarrow$}
&ArtGS~\cite{liu2025building} &195.80 &212.40 &185.60 &178.50 &188.20 &192.10\\
&REArtGS++~\cite{wu2025reartgs++} &138.20 &142.60 &135.40 &128.50 &132.10 &135.36\\
&SPAGS (Ours)&\textbf{22.40} &\textbf{26.50} &\textbf{20.80} &\textbf{19.50} &\textbf{18.60} &\textbf{21.56}\\
\midrule
\multirow{3}{*}{\shortstack{CD}$\downarrow$}
&ArtGS~\cite{liu2025building} &77.95 &85.35 &64.05 &56.95 &68.20 &70.50\\
&REArtGS++~\cite{wu2025reartgs++}  &30.30 &50.55 &45.30 &40.55 &24.45 &38.23\\
&SPAGS (Ours) &\textbf{9.45} &\textbf{22.35} &\textbf{18.10} &\textbf{17.80} &\textbf{6.70} &\textbf{14.88}\\
\bottomrule 
\end{tabular}
}
\end{table}







\begin{table}[h]
\centering
\small
\setlength{\tabcolsep}{1mm}
\caption{Ablation of view perception on PARIS dataset. Mean results are reported.}
\resizebox{0.7\linewidth}{!}{
\begin{tabular}{c|cccc}
\toprule

Settings   &CD $\downarrow$ &CD-m $\downarrow$ &CD-s $\downarrow$ \\
\hline
Random Sampling & 14.89 &40.67 &78.01\\
Pre-defined  & 10.23 &  32.34& 53.02\\
Optimal Perception &\textbf{9.60} &\textbf{25.87} &\textbf{31.02}\\
\bottomrule
\end{tabular}
}
\label{tab:ablation_view}
\vspace{-1em}
\end{table}

\subsection{Ablation Study}
\textbf{Ablation on View Perception.}
We conduct the ablation study of view perception on PARIS dataset. Quantitative results are shown in Table.~\ref{tab:ablation_view}. ``Random Sampling" denotes selecting 4-view images randomly from candidates and we take the average results over 10 trials. ``Pre-defined" refers to manually choosing promising 4-view images covering 360 degrees around the objects.

Using optimal view perception achieves the best whole surface and part-level reconstruction performance across all metrics, indicating that the Gaussian information fields capture the maximum information potential gain. 

\begin{table}[h]
\centering
\small
\setlength{\tabcolsep}{1mm}
\caption{Ablation of the number of input views on PARIS dataset. Mean results are reported.}
\begin{tabular}{c|cccc}
\toprule

Input Views   &CD $\downarrow$ &CD-m $\downarrow$ &CD-s $\downarrow$ \\
\hline
16 &\textbf{5.32}  &\textbf{16.71} &\textbf{10.25}\\
8  & 8.61 &23.41  &41.54 \\
4 &9.60 &25.87 &31.02\\
3  &15.40  & 41.23 &60.37\\
2  & 72.08 & 118.63 & 168.37\\

\bottomrule
\end{tabular}

\label{tab:ablation_view_number}
\end{table}

\textbf{Ablation of The Number of Input Views}.
To verify the performance of our method for different numbers of input views, we conduct ablation experiments on the PARIS dataset. Note that all input views are perceived by the optimal view perception method. The experimental results in Table.~\ref{tab:ablation_view_number} prove that our method can enhance the performance of mesh reconstruction by using more input views. Moreover, we observe that when the input views are less than 4, the reconstruction quality suffers a sharp decline. This is because the information provided by too few views is insufficient, and high-fidelity mesh reconstruction through the insufficient image supervision poses a significant challenge.

\begin{table}[h]
\centering
\caption{Quantitative results for  mesh reconstruction on real-world data.}
\resizebox{\linewidth}{!}{
\begin{tabular}{c|c|ccccccc}
\toprule
Metrics   &Method      & Stapler  & USB &  Scissor & Knife & Drawer & Bin  & Mean \\
\hline

\multirow{4}{*}{CD $\downarrow$}
& CoherentGS~\cite{paliwal2024coherentgs}  &30.65 &10.92 &13.12 &8.76 &104.02 &64.84 &38.72\\
& Sparse2DGS~\cite{Wu_2025_CVPR} &34.07 &7.63 &15.24 &3.21 &84.15 &65.38 &34.95\\
& GaussianObject~\cite{yang2024gaussianobject} &11.24 &3.09 &\textbf{5.23} &4.76 &21.54  &15.30 &10.19 \\
& SPAGS (Ours) & \textbf{3.13} & \textbf{2.34} & 5.76 & \textbf{2.82} & \textbf{16.36} & \textbf{4.00} & \textbf{5.74}\\

\hline
\multirow{4}{*}{F1 $\uparrow$}
& CoherentGS~\cite{paliwal2024coherentgs} &0.11 &0.14 &0.02 &0.16 &0.01 &0.02 &0.08 \\
& Sparse2DGS~\cite{Wu_2025_CVPR} &0.14 &0.20 &0.02 &0.21 &\textbf{0.03} &0.03 &0.11\\
& GaussianObject~\cite{yang2024gaussianobject} &0.16 &\textbf{0.25} &0.05 &0.22 &0.01 &\textbf{0.07} & 0.13 \\
& SPAGS (Ours) &  \textbf{0.18} & 0.24 & \textbf{0.10} & \textbf{0.27} & \textbf{0.03} & \textbf{0.07} & \textbf{0.15}\\

\hline
\multirow{4}{*}{EMD $\downarrow$}
& CoherentGS~\cite{paliwal2024coherentgs} &0.25 &0.23 &0.35 &0.15 &0.64 &0.59 &0.37\\
& Sparse2DGS~\cite{Wu_2025_CVPR} &0.16 &0.18 &0.23 &0.18 &0.58 &0.50 &0.31\\
& GaussianObject~\cite{yang2024gaussianobject} & 0.09 &0.11 &0.16 &0.14 &0.23 &0.27 &0.17 \\
& SPAGS (Ours) &  \textbf{0.06}  & \textbf{0.07} & \textbf{0.10} & \textbf{0.07} & \textbf{0.13} & \textbf{0.08} & \textbf{0.09} \\

\bottomrule
\end{tabular}
}
\label{tab:real_world}
\end{table}

\textbf{Ablation of Key Components.}
We conduct ablation of proposed core components on PARIS dataset. In the ``w/o planar Gaussians" setting, we employ the depth rendering pipeline from GaussianObject. In ``w/o pseudo labels" setting, we omit the pseudo labels generated from reliable regions.  The experimental results in Table.~\ref{tab:key_ablation} demonstrate that each  component makes a significant improvement to both whole surface and part-level reconstruction performance, especially for the planar Gaussians and the refinement. This confirms the effectiveness of proposed key components, and the superiority of planar Gaussians as well as the refinement via few-shot diffusion.

\begin{table}[h]
\centering
\small
\setlength{\tabcolsep}{1mm}
\caption{Ablation of key components on PARIS dataset. Mean results are reported.}
\scalebox{0.9}{
\begin{tabular}{c|cccc}
\toprule

Settings   &CD $\downarrow$ &CD-m $\downarrow$ &CD-s $\downarrow$ \\
\hline

w/o depth smooth regularization &11.32 &31.84 & 54.81 \\
w/o depth regularization  &15.84 & 34.20 &     59.32 \\

w/o pseudo labels &10.27 &30.58 & 51.08\\
w/o view regularization &10.18 &30.08 &48.63\\
w/o refinement &75.63 &81.05 &90.47\\

w/o planar Gaussians &30.98 &61.52 &76.25 \\
w/ all &\textbf{9.60} &\textbf{25.87} &\textbf{31.02}\\

\bottomrule
\end{tabular}
}

\label{tab:key_ablation}
\end{table}


\subsection{Training Time Analysis}
On a single RTX 4090 GPU, our method takes about 45 minutes for the object-level reconstruction, which is comparable to GaussianObject. Concretely, the coarse-to-fine optimization takes about 40 minutes, making up most of the training time. The optimal viewpoint perception only requires about few minutes and representation initialization takes about 1 minute. The articulated modeling requires only 2 minutes approximately.

\begin{figure}[h]
    \centering
    \vspace{-10pt}
    \includegraphics[width=\linewidth]{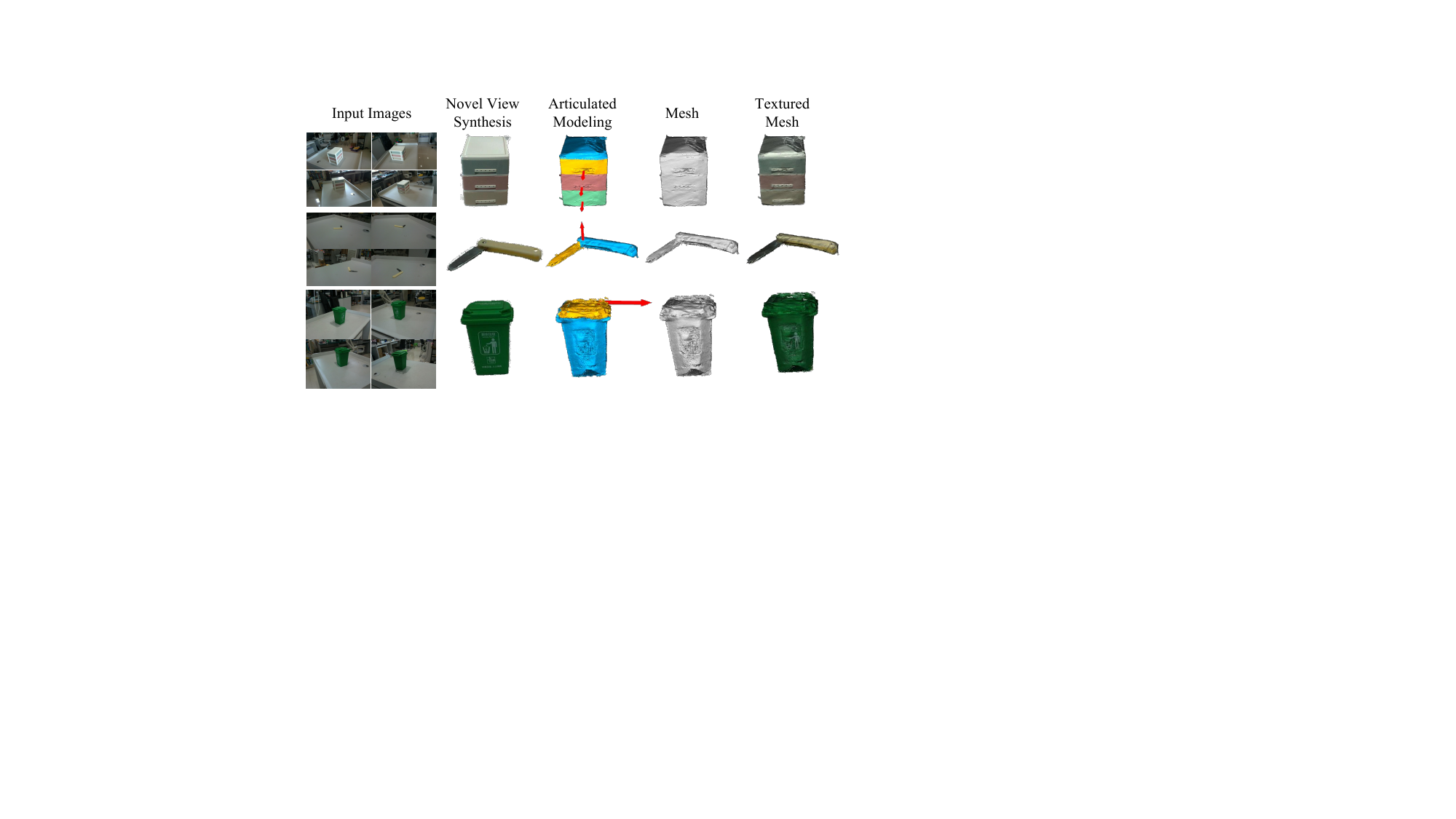}

    \caption{Qualitative results of our method for real-world articulated objects.}
    \label{fig:real_world}
    \vspace{-10pt}
\end{figure}

\subsection{Generalization to the Real World}
To further evaluate our generalization to real-world objects, we reconstruct objects by capturing sparse images of indoor articulated  objects under a static state. Since methods like ArtGS and REArtGS++ are inapplicable using single-state observations, we conduct quantitative comparisons  for whole meshes against SOTA sparse reconstruction baselines in Tab.~\ref{tab:real_world}, and provide qualitative results of articulated modeling in Fig.~\ref{fig:real_world}.

As shown in Tab.~\ref{tab:real_world}, our method significantly outperforms existing approaches in mean results across all metrics and demonstrates strong generalization on real-world data. As shown in Fig.~\ref{fig:real_world}, our method enables high-fidelity reconstruction for unseen real-world articulated objects with two or more parts.

\subsection{Conclusion and Limitation}
In this paper, we propose a category-agnostic articulated object reconstruction via planar Gaussians, which requires only single-stage sparse-view RGB images. We introduce an optimal sparse view perception approach through Gaussian information fields and perform a coarse-to-fine optimization as well as articulated modeling for planar Gaussians. Extensive experiments  demonstrate our superiority compared to existing approaches.


Despite its effectiveness, our method exhibits limitations when handling real-world articulated objects with transparent materials or minute geometric structures. Standard Gaussian representations struggle to capture complex optical phenomena like refraction, while extremely small parts suffer from insufficient pixel-level constraints in sparse settings. Moving forward, we intend to integrate physically based rendering (PBR)~\cite{transparentgs} into our pipeline to explicitly model transparent and specular surfaces. Furthermore, we aim to incorporate neural super-resolution techniques~\cite{xie2024supergs} to recover high-frequency details from low-resolution inputs, enabling robust reconstruction of micro-articulated structures.



\section*{Acknowledgements}
This work was supported by National Natural Science Foundation of China under Grant 62302143.

\nocite{langley00}

\bibliography{main}
\bibliographystyle{IEEEtran}


 




\vfill

\end{document}